\documentclass[runningheads]{llncs}
\usepackage{makeidx}
\usepackage{graphicx}
\usepackage{amsmath,amssymb} 
\usepackage{color}
\usepackage{booktabs}
\usepackage{bbm}

\newcommand{\etal}{et al.}
\newcommand{\ie}{\emph{i.e.}}
\newcommand{\eg}{\emph{e.g.}}

\begin{document}
	\pagestyle{headings}
	\mainmatter

	\def\GCPR20SubNumber{37}

	\title{Boosting Generalization in Bio-Signal Classification by Learning the Phase-Amplitude Coupling}  
    
	\titlerunning{Boosting Generalization in Bio-Signal Classification}
	\authorrunning{A. Lemkhenter et P. Favaro}
	\author{Abdelhak Lemkhenter \orcidID{0000-0001-6543-4772} \and Paolo Favaro\orcidID{0000-0003-3546-8247}}
	\institute{Department of Computer Science, University of Bern \{abdelhak.lemkhenter,paolo.favaro@inf.unibe.ch\}}

	\maketitle

    \begin{abstract}
	Various hand-crafted feature representations of bio-signals rely primarily on the amplitude or power of the signal in specific frequency bands. The phase component is often discarded as it is more sample specific, and thus more sensitive to noise, than the amplitude. However, in general, the phase component also carries information relevant to the underlying biological processes. In fact, in this paper we show the benefits of learning the coupling of both phase and amplitude components of a bio-signal. We do so by introducing a novel self-supervised learning task, which we call \emph{phase-swap}, that detects if bio-signals have been obtained by merging the amplitude and phase from different sources. We show in our evaluation that neural networks trained on this task generalize better across subjects and recording sessions than their fully supervised counterpart.
	\end{abstract}
	
	\section{Introduction}
	
Bio-signals, such as Electroencephalograms and Electrocardiograms, are multivariate time-series generated by biological processes that can be used to assess seizures, sleep disorders, head injuries, memory problems, heart diseases, just to name a few \cite{nait2009advanced}. Although clinicians can successfully learn to correctly interpret such bio-signals, their protocols cannot be directly converted into a set of numerical rules yielding a comparable assessment performance.
Currently, the most effective way to transfer this expertise into an automated system is to gather a large number of examples of bio-signals with the corresponding labeling provided by a clinician, and to use them to train a deep neural network. However, collecting such labeling is expensive and time-consuming. In contrast, bio-signals without labeling are more readily available in large numbers.

Recently, self-supervised learning (SelfSL) techniques have been proposed to limit the amount of required labeled data. These techniques define a so-called \emph{pretext} task that can be used to train a neural network in a supervised manner on data without manual labeling. The pretext task is an artificial problem, where a model is trained to output what transformation was applied to the data. For instance, a model could be trained to output the probability that a time-series had been time-reversed \cite{wei2018learning}. This step is often called pre-training and it can be carried out on large data sets as no manual labeling is required. The training of the pre-trained neural network then continues with a small learning rate on the small target data set, where labels are available. This second step is called  \emph{fine-tuning}, and it yields a substantial boost in performance \cite{noroozi2016unsupervised}. Thus, SelfSL can be used to automatically learn physiologically relevant features from unlabelled bio-signals and improve classification performance.

SelfSL is most effective if the pretext task focuses on features that are relevant to the target task. Typical features work with the amplitude or the power of the bio-signals, but as shown in the literature, the phase carries information about the underlining biological processes \cite{busch2009phase,ng2013eeg,lopez2019dynamic}. Thus, in this paper, we propose a pretext task to learn the coupling between the amplitude and the phase of the bio-signals, which we call \emph{phase swap} (PS). The objective is to predict whether the phase of the Fourier transform of a multivariate physiological time-series segment was swapped with the phase of another segment. 

We show that features learned through this task help classification tasks generalize better, regardless of the neural network architecture.

Our contributions are summarized as follows
\begin{itemize}
    \item We introduce phase swap, a novel self-supervised learning task to detect the coupling between the phase and the magnitude of physiological time-series;
    \item With phase swap, we demonstrate experimentally the importance of incorporating the phase in bio-signal classification;
    \item We show that the learned representation generalizes better than current state of the art methods to new subjects and to new recording sessions;
    \item We evaluate the method on four different data sets and analyze the effect of various hyper-parameters and of the amount of available labeled data on the learned representations.
\end{itemize}

    \section{Related Work}
    
    \noindent\textbf{Self-supervised Learning.} Self-supervised learning refers to the practice of pre-training deep learning architectures on user-defined pretext tasks. This can be done on large volumes of unlabeled data since the annotations can be automatically generated for these tasks. This is a common practice in the Natural Language Processing literature. Examples of such works include Word2Vec \cite{mikolov2013efficient}, where the task is to predict a word from its context, and BERT \cite{devlin2018bert}, where the model is pretrained as a masked language model and on the task of detecting consecutive sentences. The self-supervision framework has also been gaining popularity in Computer Vision.
    Pretext tasks such as solving a jigsaw puzzle \cite{noroozi2016unsupervised}, predicting image rotations \cite{gidaris2018unsupervised} and detecting local inpainting \cite{jenni2020steering} have been shown to be able to learn useful data representations for downstream tasks.
    Recent work explores the potential of self-supervised learning for EEG signals \cite{banville2019self} and time series in general \cite{jawed2020self}. In \cite{banville2019self}, the focus is on long-term/global tasks such as determining whether two given windows are nearby temporally or not. \\
    \noindent\textbf{Deep Learning for Bio-signals.} Bio-signals include a variety of physiological measures across time such as: Electroencephalogram (EEG), Electrocardiogram (ECG), Electromyogram (EMG), Electrooculography (EOG), etc.
    These signals are used by clinicians in various applications, such as sleep scoring \cite{mourtazaev1995age} or seizure detection \cite{shoeb2009application}.
    Similarly to many other fields, bio-signals analysis has also seen the rise in popularity of deep learning methods for both classification \cite{humayun2019end} and representation learning \cite{banville2019self}.The literature review \cite{roy2019deep} showcases the application of deep learning methods to various EEG classification problems such as brain computer interfaces, emotion recognition and seizure detection. The work by Banville \etal~\cite{banville2019self} leverages self-supervised tasks based on the relative temporal positioning of pairs/triplets of EEG segments to learn a useful representation for a downstream sleep staging application.\\
    \noindent\textbf{Phase Analysis.} The phase component of bio-signals has been analyzed before. Busch \etal~\cite{busch2009phase} show a link between the phase of the EEG oscillations, in the alpha (8-12Hz) and theta (4-8Hz) frequency bands, and the subjects' ability to perceive the flash of a light. The phase of the EEG signal is also shown to be more discriminative for determining firing patterns of neurons in response to certain types of stimuli \cite{ng2013eeg}.
    More recent work, such as \cite{lopez2019dynamic}, highlights the potential link between the phase of the different EEG frequency bands and cognition during proactive control of task switching.

    \begin{figure}[t]
        \centering
        \includegraphics[width=.75\textwidth,trim=0 1.95cm 0 2cm,clip]{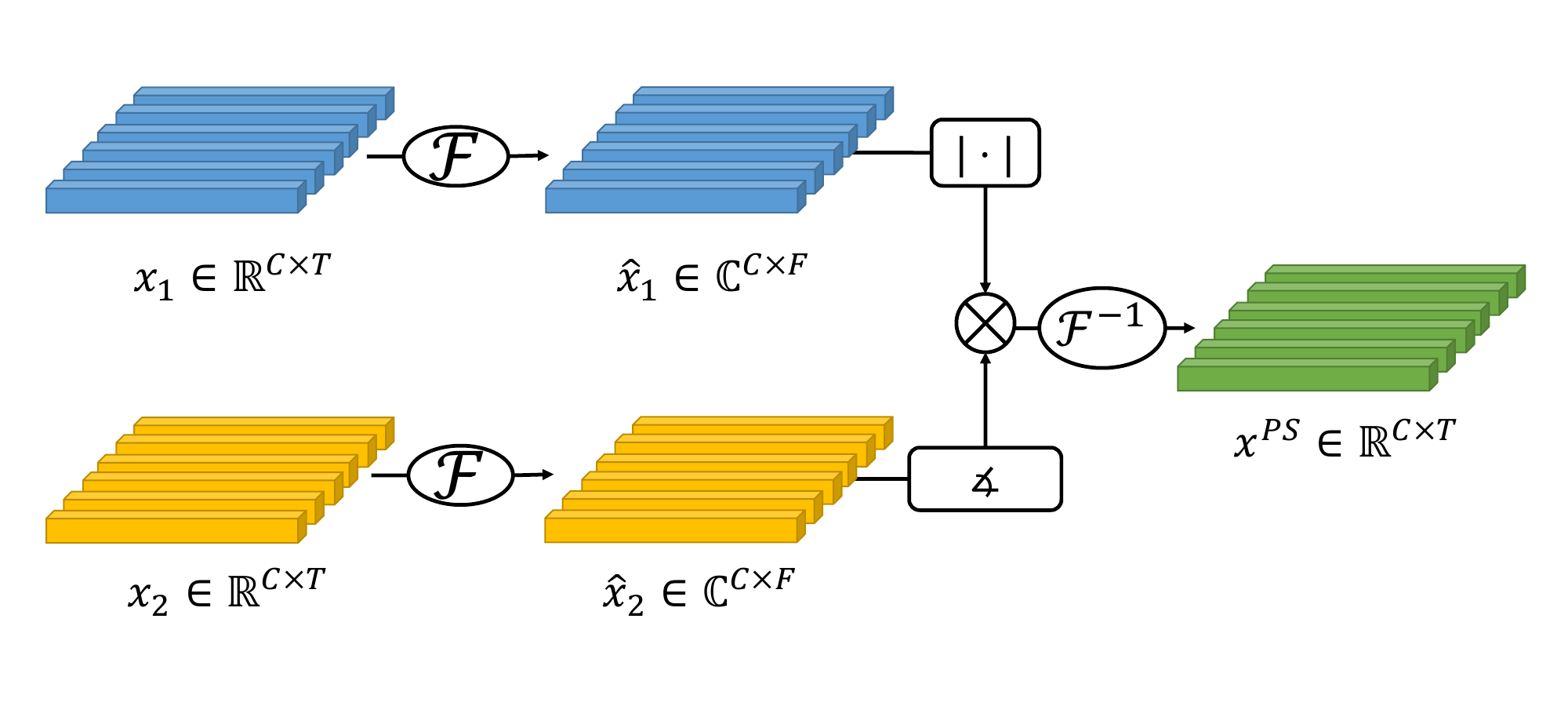}
        \caption{Illustration of the phase-swap operator $\Phi$. The operator takes two signals as input and then combines the amplitude of the first signal with the phase of the second signal in the output.}
        \label{fig:phaseswap_diag}
    \end{figure}

    \section{Learning to Detect the Phase-Amplitude Coupling}
    \label{sec:phaseswap}

   In this section, we define the \emph{phase swap} operator and the corresponding SelfSL task, and present the losses used for pre-training and fine-tuning.

    Let $D^W_{i,j} = \{(x^{i, j, k}, y^{i, j, k})\}_{k=1}^N$ be the set of samples associated with the i-th subject during the j-th recording session. Each sample $x^{i, j, k} \in \mathbf{R}^{C \times W}$ is a multivariate physiological time-series window where $C$ and $W$ are the number of channels and the window size respectively. $y^{i, j, k}$ is the class of the k-th sample.
    Let $\mathcal{F}$ and $\mathcal{F}^{-1}$ be the Discrete Fourier Transform operator and its inverse, respectively.  
    These operators will be applied to a given vector $x$ extracted from the bio-signals. In the case of multivariate signals, we apply these operators channel-wise. 

    For the sake of clarity, we provide the definitions of the absolute value and the phase element-wise operators. Let $z\in \mathbf{C}$, where $\mathbf{C}$ denotes the set of complex numbers. Then, the absolute value, or \emph{magnitude}, of $z$ is denoted $|z|$ and the phase of $z$ is denoted $\measuredangle z$. With such definitions, we have the trivial identity $z = |z|\measuredangle z$.

    \begin{figure}[t]
        \centering
        \includegraphics[width=\textwidth,trim=0 .2cm 0 .2cm,clip]{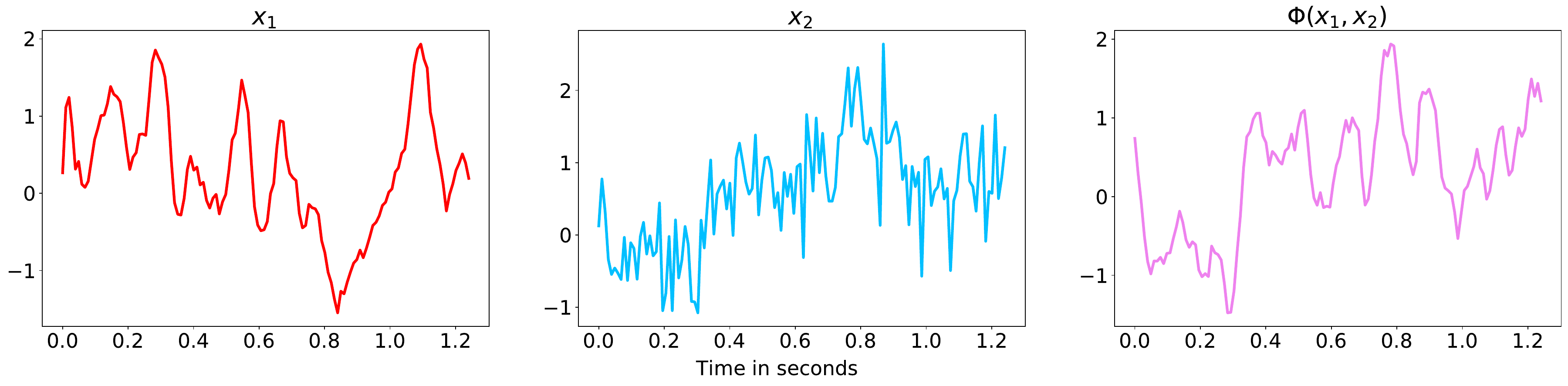}
        \caption{Illustration of the PS operator on a pair of 1.25 seconds segments taken from the Fpz-Cz channel in the SC data set \cite{mourtazaev1995age}. The original signals are $x_1$ and $x_2$.}
        \label{fig:phaseswap}
    \end{figure}

    Given two samples $x^{i, j, k}$, $x^{i, j, k'} \in D^W_{i, j}$, the \emph{phase swap} (PS) operator $\Phi$ is 
    \begin{align}
    \label{eq:swap}
    \textstyle
        \Phi \left(x^{i, j, k}, x^{i, j, k'}\right) \doteq \mathcal{F}^{-1} \left[ \left|\mathcal{F}\left(x^{i, j, k}\right)\right| \odot \measuredangle \mathcal{F}\left(x^{i, j, k'}\right)\right] = x^{i, j, k}_{swap},
    \end{align}
    where $\odot$ is the element-wise multiplication (see Fig.~\ref{fig:phaseswap_diag}).
    Note that the energy per frequency is the same for both $x^{i,k}_{swap}$ and $x^{i,k}$ and that only the phase, \ie, the synchronization between the different frequencies, changes. Examples of phase swapping between different pairs of signals are shown in Fig.~\ref{fig:phaseswap}. 
    
    Notice how the shape of the oscillations change drastically when the PS operator is applied and no trivial shared patterns seem to emerge.
    
    The PS pretext task is defined as a binary classification problem. A sample belongs to the positive class if it is transformed using the PS operator, otherwise it belongs to the negative class. In all our experiments, both inputs to the PS operator are sampled from the same patient during the same recording session. 
    Because the phase is decoupled from the amplitude of white noise, our model has no incentive to detect noise patterns. On the contrary, it will be encouraged to focus on the structural patterns in the signal in order to detect whether the phase and magnitude of the segment are coupled or not. 
    
    \label{sec:archi}
    
    \begin{figure}[t]
        \centering
        \includegraphics[width=1\textwidth,trim=0 2.1cm 0cm 0cm,clip]{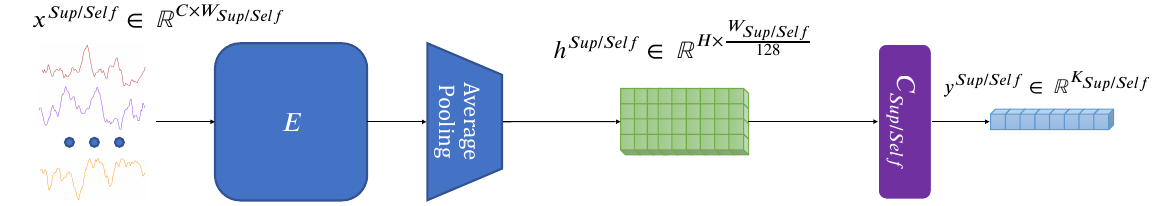}
        \caption{Training with either the self-supervised or the supervised learning task.}
        \label{fig:overview}
    \end{figure}
    
    We use the FCN architecture proposed by Wang \etal~\cite{wang2017time} as our core neural network model $E:  \mathbf{R}^{C\times W} \to \mathbf{R}^{H \times W/128}$. It consists of 3 convolutions blocks using a Batch Normalization layer \cite{ioffe2015batch} and a ReLU activation followed by a pooling layer. The output of $E$ is then flattened and fed to two Softmax layers $C_{Self}$ and $C_{Sup}$, which are trained on the self-supervised and supervised tasks respectively. 
   
    Instead of a global pooling layer, we use an average pooling layer with a stride of 128. This allows us to keep the  number of weights of the supervised network $C_{Sup} \circ E$ constant when the self-supervised task is defined on a different window size. 
    The overall framework is illustrated in Fig.~\ref{fig:overview}. Note that the encoder network $E$ is the same for both tasks.
    
    The loss function for training on the SelfSL task is the cross-entropy
    \begin{align}
    \textstyle
    \mathcal{L}_{Self}\left(y^{Self},  E, C_{Self}\right) = - \frac{1}{N} \sum_{i=1}^N \sum_{k=1}^{K_{Self}} y^{Self}_{i,k} \log \left(C_{Self} \circ E(x_i)\right)_k,
    \label{eq:loss_ssl} 
    \end{align}
    
    where $y^{Self}_{i,k}$ and $(C_{Self} \circ E(x_i))_k$ are the one-hot representations of the true SelfSL pretext label and the predicted probability vector respectively. 
    We optimize eq.~\eqref{eq:loss_ssl} with respect to the parameters of both $E$ and $C_{Self}$.
    Similarly, we define the loss function for the (supervised) fine-tuning as the cross-entropy
    \begin{align}
    \textstyle
    \mathcal{L}_{Sup}\left(y^{Sup},  E, C_{Sup}\right) = - \frac{1}{N} \sum_{i=1}^N \sum_{k=1}^{K_{Sup}} y^{Sup}_{i,k} \log \left(C_{Sup} \circ E(x_i)\right)_k,
    \label{eq:loss_sup} 
    \end{align}    
    where $y^{Sup}_{i,k}$ denotes the label for the target task. The $y^{Sup/Self}_{i,k}$
    vectors are in $\mathbf{R}^{N \times K_{Sup/Self}}$, where $N$ and $K_{Sup/Self}$ are the number of samples and classes respectively. In the fine-tuning, $E$ is initialized with the parameters obtained from the optimization of eq.~\eqref{eq:loss_ssl} and $C_{Sup}$ with random weights, and then they are both updated to optimize eq.~\eqref{eq:loss_sup}, but with a small learning rate.
    
    \section{Experiments}
    
    \subsection{Data Sets}
    
    In our experiments, we use the Expanded SleepEDF \cite{mourtazaev1995age,kemp2000analysis,goldberger2000physiobank}, the CHB-MIT \cite{shoeb2009application} and ISRUC-Sleep \cite{khalighi2016isruc} data sets as they contain recordings from multiple patients. This allows us to study the generalization capabilities of the learned feature representation to new recording sessions and new patients.
    The Expanded SleepEDF database contains two different sleep scoring data sets
    \begin{itemize}
        \item Sleep Cassette Study (SC) \cite{mourtazaev1995age}: Collected between 1987 and 1991 in order to study the effect of age on sleep. It includes 78 patients with 2 recording sessions each (3 recording sessions were lost due to hardware failure). 
        
        \item Sleep Telemetry Study (ST) \cite{kemp2000analysis}: Collected in 1994 as part of a study of the effect of Temazepam on sleep in 22 different patients with 2 recordings sessions each. 
    \end{itemize}
    Both data sets define sleep scoring as a 5-way classification problem. The 5 classes in question are the sleep stages: Wake, NREM 1, NREM 2, NREM 3/4, REM. The NREM 3 and 4 are merged into one class due to their small number of samples (these two classes are often combined together in sleep studies). \\
    The third data set we use in our experiments is the CHB-MIT data set \cite{shoeb2009application} recorded at the Children’s Hospital Boston from pediatric patients with intractable seizures. It includes multiples recording files across 22 different patients. We retain the 18 EEG channels that are common to all recording files.
    The sampling rate for all channels is 256Hz. The target task defined on this data set is predicting whether a given segment is a seizure event or not, \ie, a binary classification problem. 
    For all the data sets, the international 10-20 system \cite{malmivuo1995bioelectromagnetism} was adopted for the choice of the positioning of the EEG electrodes. 
    
    The last data set we use is ISRUC-Sleep \cite{khalighi2016isruc}, for sleep scoring as a 4-way classification problem. We use the 14 channels extracted in the Matlab version of the data set. This data set consists of three subgroups: subgroups I and II contain respectively recordings from 100 and 8 subjects with sleep disorders, whereas subgroup III contains recordings from 10 healthy subjects. This allows us to test the generalization from diagnosed subjects to healthy ones.
    
    For the SC, ST and ISRUC-sleep data sets we resample the signals to 102.4Hz. This resampling allows us to simplify the neural network architectures we use, because in this case most window sizes can be represented by a power of 2, \eg, a window of 2.5sec corresponds to 256 samples.
    We normalize each channel per recording file in all data sets to have zero mean and a standard deviation of one.
    
    \subsection{Training Procedures and Models}
    
    In the supervised baseline (respectively, self-supervised pre-training), we train the randomly initialized model $C_{Sup} \circ E$ (respectively, $C_{Self} \circ E$) on the labeled data set for 10 (respectively, 5) epochs using the Adam optimizer \cite{KingmaB14Adam} with a learning rate of $10^{-3}$ and $\beta = (0.9, 0.999)$. We balance the classes present in the data set using resampling (no need to balance classes in the self-supervised learning task).
    In fine-tuning, we initialize $E$'s weights with those obtained from the SelfSL training and then train $C_{Sup} \circ E$ on the labeled data set for 10 epochs using the Adam optimizer \cite{KingmaB14Adam}, but with a learning rate of $10^{-4}$ and $\beta = (0.9, 0.999)$. As in the fully supervised training, we also balance the classes using re-sampling.
    In all training cases, we use a default batch size of 128. 

    \label{sec:models}
    
    We evaluate our self-supervised framework using the following models
    \begin{itemize}
        \item\textbf{PhaseSwap}: The model is pre-trained on the self-supervised task and fine-tuned on the labeled data;
        \item\textbf{Supervised}: The model is trained solely in a supervised fashion;
        \item\textbf{Random}: $C_{Sup}$ is trained on top of a frozen randomly initialized $E$;
        \item\textbf{PSFrozen}: We train $C_{Sup}$ on top of the frozen weights of the model $E$ pre-trained on the self-supervised task.
    \end{itemize}

    \subsection{Evaluation Procedures}
    
    We evaluate our models on train/validation/test splits in our experiments. In total we use at most 4 sets, which we refer to as the training set, the Validation Set, the Test set A and the Test set B.
    The validation set and test set A and the training set share the same patient identities, while B contains recordings from other patients. The validation set and test set A use distinct recording sessions. Validation Set and the training set share the same patient identities and recording sessions with a $75\%$ (for the training set) and $25\%$ (Validation Set) split.
    We use each test set for the following purposes
    \begin{itemize}
        \item \textbf{Validation Set}: this set serves as a validation set;
        \item \textbf{Test set A}: this set allows us to evaluate the generalization error on new recording sessions for patients observed during training;
        \item \textbf{Test set B}: this set allows us to evaluate the generalization error on new recording sessions for patients not observed during training.
    \end{itemize}
    We use the same set of recordings and patients for both the training of the self-supervised and supervised tasks.
    For the ST, SC and ISRUC data sets we use class re-balancing only during the supervised fine-tuning. However, for the CHB-MIT data set, the class imbalance is much more extreme: The data set consists of less than $0.4\%$ positive samples. Because of that, we under-sample the majority class both during the self-supervised and supervised training. This prevents the self-supervised features from completely ignoring the positive class. Unless specified otherwise, we use $W_{Self}=5 sec$ and $W_{Sup}=30 sec$ for the ISRUC, ST and SC data sets, $W_{Self}= 2sec$ and $W_{Sup} = 10 sec$ for the CHB-MIT data set, where $W_{Self}$ and $W_{Sup}$ are the window size for the self-supervised and supervised training respectively. For the ISRUC, ST and SC data sets, the choice of $W_{Sup}$ corresponds to the granularity of the provided labels. For the CHB-MIT data set, although labels are provided at a rate of 1Hz, the literature in neuroscience usually defines a minimal duration of around 10sec for an epileptic event in humans \cite{fisher2014can}, which motivates our choice of $W_{Sup} = 10 sec$. \\
    \noindent\textbf{Evaluation Metric.} As an evaluation metric, we use the balanced accuracy 
    \begin{equation}
    \label{eq:balanced}
        Acc^{Balanced} (y, \hat{y}) = \frac{1}{K} \sum_{k=1}^K \frac{ \sum_{i=1}^N \hat{y}_{i,k} y_{i,k}
         }{\sum_{i=1}^N y_{i,k}},
    \end{equation}    
    which is defined as the average of the recall values per class, where $K$, $N$, $y$ and $\hat{y}$ are respectively the number of classes, the number of samples, the one-hot representation of true labels and the predicted labels. 
    
    \subsection{Generalization on the Sleep Cassette Data Set}
    We explore the generalization of the self-supervised trained model by varying the number of different patients used in the training set for the SC data set. The $r_{train}$ is the percentage of patient identities used for training, in Validation Set and in Test set A. In Table~\ref{tab:generalization}, we report the balanced accuracy on all test sets for various values of $r_{train}$. The self-supervised training was done using a window size of $W_{Self}=5sec$. We observe that the \textbf{PhaseSwap} model performs the best for all values of $r_{train}$. We also observe that the performance gap between the \textbf{PhaseSwap} and \textbf{Supervised} models is narrower for larger values for $r_{train}$. This is to be expected since including more identities in the training set allows the \textbf{Supervised} model to generalize better. For $r_{train} = 100\%^*$ we use all recording sessions across all identities for the training set and in the Validation Set (since all identities and sessions are used, the Test sets A and B are empty). The results obtained for this setting show that there is still a slight benefit with the \textbf{PhaseSwap} pre-training even when labels are available for most of the data. 
    
    \begin{table}[t]
    \centering
    \caption{Comparison of the performance of the \textbf{PhaseSwap} model on the SC data set for various values of $r_{train}$. For $r_{train} = 100\%^*$ we use all available recordings for the training and the Validation sets. Results with different $r_{train}$ are not comparable.}
    \label{tab:generalization}
    \begin{tabular}{c@{\hspace{2em}}c@{\hspace{2em}}r@{\hspace{2em}}r@{\hspace{2em}}r}
    \toprule
    $r_{train}$ & Experiment & Validation Set      & Test set A      & Test set B      \\ \midrule
    20\%        & \textbf{PhaseSwap}  & \textbf{84.3\%} & \textbf{72.0\%}   & \textbf{69.6\%} \\
    20\%        & \textbf{Supervised} & 79.4\%          & 67.9\%          & 66.0\%            \\ \hline
    50\%        & \textbf{PhaseSwap}  & \textbf{84.9\%} & \textbf{75.1\%} & \textbf{73.3\%} \\
    50\%        & \textbf{Supervised} & 81.9\%          & 71.7\%          & 69.4\%          \\ \hline
    75\%        & \textbf{PhaseSwap}  & \textbf{84.9\%} & \textbf{77.6\%} & \textbf{76.1\%} \\
    75\%        & \textbf{Supervised} & 81.6\%          & 73.7\%          & 72.8\%          \\ \hline
    100\%*      & \textbf{PhaseSwap}  & \textbf{84.3\%} & -               & -               \\
    100\%*      & \textbf{Supervised} & 83.5\%          & -               & -               \\ \bottomrule
    \end{tabular}
    \end{table}
    
    \subsection{Generalization on the ISRUC-Sleep Data Set}
    Using the ISRUC-Sleep data set \cite{khalighi2016isruc}, we aim to evaluate the performance of the PhaseSwap model on healthy subjects when it was trained on subjects with sleep disorders. For the self-supervised training, we use $W_{Self} = 5 sec$. The results are reported in Table~\ref{tab:isruc}. Note that we combined the recordings of subgroup II and the ones not used for the training from subgroup I into a single test set since they are from subjects with sleep disorders. We observe that for both experiments, $r_{train} = 25 \%$ and $r_{train} = 50 \%$, the PhaseSwap model outperforms the supervised baseline for both test sets. Notably, the performance gap on subgroup III is larger than $10 \%$. This can be explained by the fact that sleep disorders can drastically change the sleep structure of the affected subjects, which in turn leads the supervised baseline to learn features that are specific to the disorders/subjects present in the training set.
    
    \begin{table}[t]
    \centering
    \caption{Comparison of the performance of the \textbf{PhaseSwap} model on the ISRUC-Sleep data set for various values of $r_{train}$.}
    \label{tab:isruc}
    \begin{tabular}{c@{\hspace{2em}}c@{\hspace{2em}}r@{\hspace{2em}}r@{\hspace{2em}}r}
    \toprule
    $r_{train}$ & Model      & Validation set   & \begin{tabular}[c]{@{}c@{}}Test set B \\ (subgroup I + II)\end{tabular} & \begin{tabular}[c]{@{}c@{}}Test set B\\ (subgroup III)\end{tabular} \\ \midrule
    25\%          & PhaseSwap  & 75.8\%           & \textbf{67.3\%}  & \textbf{62.8\%}                                                    \\
    25\%          & Supervised & \textbf{75.9\%}  & 63.1\%           & 47.9\%                                                             \\ \hline
    50\%          & PhaseSwap  & \textbf{76.3\%} & 68.2\% & \textbf{67.1\%}                                                    \\
    50\%          & Supervised & 75.5\%          & \textbf{68.3}\%          & 57.3\%                                                             \\ \bottomrule
    \end{tabular}
    \end{table}

    \subsection{Comparison to the Relative Positioning Task}
    
    The Relative Positioning (RP) task was introduced by Banville \etal~\cite{banville2019self} as a self-supervised learning method for EEG signals, which we briefly recall here.
    Given $x_t$ and $x_{t'}$, two samples with a window size $W$ and starting points $t$ and $t'$ respectively, the RP task defines the following labels

    $C_{Self}(|h_t - h_{t'}|) = \mathbbm{1}\left(|t- t'| \leq \tau_{pos}\right)-\mathbbm{1}\left(|t- t'| > \tau_{neg}\right)$,
    where $h_t = E(x_t)$, $h_{t'} = E(x_{t'})$, $\mathbbm{1}(\cdot)$ is the indicator function, and $\tau_{pos}$ and $\tau_{neg}$ are predefined quantities. Pairs that yield $C_{Self}=0$ are discarded. $|\cdot|$ denotes the element-wise absolute value operator.
    
    Next, we compare our self-supervised task to the RP task \cite{banville2019self}. For both settings, we use $W_{Self} = 5 sec$ and $r_{train}=20\%$. For the the RP task we choose $\tau_{pos}=\tau_{neg}= 12 \times W_{Self}$. We report the balanced accuracy for all test sets on the SC data set in Table~\ref{tab:comparison}. We observe that our self-supervised task outperforms the RP task. This means that the features learned through the PS task allow the model to perform better on unseen data. 
    
    \begin{table}[t]
    \centering
    \caption{Comparison between the PS and RP pre-training on the SC data set.}
    \label{tab:comparison}
    \begin{tabular}{c@{\hspace{1em}}c@{\hspace{1em}}c@{\hspace{1em}}c@{\hspace{2em}}c}
    \toprule
        Pre-training       & Validation Set      & Test set A    & Test set B      & SelfSL validation accuracy \\ \midrule
    Supervised & 79.4\%          & 67.9\%        & 66.0\%          & -                       \\
    \midrule
    PS & \textbf{84.3\%} & \textbf{72.0\%} & \textbf{69.6\%} & 86.9\%                  \\
    RP         & 80.3\%          & 66.2\%        & 65.4\%          & 56.9\%                  \\ \bottomrule
    \end{tabular}
    \end{table}
    
    \subsection{Results on the Sleep Telemetry and CHB-MIT Data Sets}
    
    In this section, we evaluate our framework on the ST and CHB-MIT data sets. For the ST data set, we use $W_{Self}= 1.25sec$, $W_{Sup}=30sec$ and $r_{train}=50\%$. For the CHB-MIT data set, we use $W_{Self}= 2sec$, $W_{Sup}= 10sec$, $r_{train}=25\%$ and 30 epochs for the supervised fine-tuning/training.
    As shown in Table~\ref{tab:other-datasets}, we observe that for the ST data set, the features learned through the PS task produce a significant improvement, especially on Test sets A and B. 
    For the CHB-MIT data set, the PS fails to provide the performance gains as observed for the previous two data sets. We believe that this is due to the fact that the PS task is too easy on this particular data set: Notice how the validation accuracy is above $99\%$. With a trivial task, self-supervised pre-training fails to learn any meaningful feature representations.
    
    In order to make the task more challenging, we introduce a new variant, which we call \textbf{PS + Masking}, where we randomly zero out all but 6 randomly selected channels for each sample during the self-supervised pre-training. The model obtained through this scheme performs the best on both sets A and B and is comparable to the \textbf{Supervised} baseline on the validation set.
    As for the reason why the PS training was trivial on this particular data set, we hypothesize that this is due to the high spatial correlation in the CHB-MIT data set samples.
    This data set contains a high number of homogeneous channels (all of them are EEG channels), which in turn result in a high spatial resolution of the brain activity. At such a spatial resolution, the oscillations due to the brain activity show a correlation both in space and time \cite{ito2005spatial}. However, our PS operator ignores the spatial aspect of the oscillations. When applied, it often corrupts the spatial coherence of the signal, which is then easier to detect than the temporal phase-amplitude incoherence. This hypothesis is supported by the fact that the random channel masking, which in turn reduces the spatial resolution during the self-supervised training, yields a lower training accuracy, \ie, it is a non-trivial task.
    
    \begin{table}[t]
    \centering
    \caption{Evaluation of the \textbf{PhaseSwap} model on the ST and CHB-MIT datasets.}
    \label{tab:other-datasets}
    \begin{tabular}{c@{\hspace{1em}}c@{\hspace{1em}}r@{\hspace{1em}}r@{\hspace{1em}}r@{\hspace{1em}}r}
    \toprule
    Dataset & Experiment        & Val. Set      & Test set A      & Test set B \quad     & SelfSL val. accuracy \\ \midrule
    ST      & \textbf{Supervised}        & 69.2\%          & 52.3\%          & 46.7\%          & -                       \\
    ST      & \textbf{PhaseSwap}         & \textbf{74.9\%} & \textbf{60.4\%} & \textbf{52.3\%} & 71.3\%                  \\ \hline
    CHB-MIT & \textbf{Supervised} & \textbf{92.6\%} & 89.5\%          & 58.0\%                          & -                       \\
    CHB-MIT & \textbf{PhaseSwap}  & 92.2\%          & 86.8\%          & 55.1\%                          & 99.8\%                  \\
    CHB-MIT & \textbf{PS+Masking} & 91.7\%          & \textbf{90.6\%} & \textbf{59.8\%}                 & 88.1\%                  \\ \bottomrule
    \end{tabular}
    \end{table}
    \begin{table}[t]
    \centering
    \caption{Comparison of the performance of the \textbf{PhaseSwap} model on the SC data set for various values of the window size $W_{Self}$.}
    \label{tab:window_size}
    \begin{tabular}{c@{\hspace{2em}}c@{\hspace{2em}}r@{\hspace{2em}}r@{\hspace{2em}}r}
    \toprule
    $W_{Self}$  & Experiment & Validation Set        & Test set A        & Test set B        \\ \midrule
    1.25sec & \textbf{PhaseSwap}  & 84.3\%          & 72.0\%          & 69.6\%          \\
    2.5sec  & \textbf{PhaseSwap}  & \textbf{84.6\%} & 71.9\%          & 70.0\%            \\
    5sec    & \textbf{PhaseSwap}  & 83.4\%          & \textbf{72.5\%} & \textbf{70.9\%} \\
    10sec   & \textbf{PhaseSwap}  & 83.6\%          & 71.6\%          & 69.9\%          \\
    30sec   & \textbf{PhaseSwap}  & 83.9\%          & 71.0\%          & 69.2\%          \\
    -       & \textbf{Supervised} & 79.4\%          & 68.1\%          & 66.1\%          \\
    \bottomrule
    \end{tabular}
    \end{table}

    \subsection{Impact of the Window Size}
    
    In this section, we analyze the effect of the window size $W_{Self}$ used for the self-supervised training on the final performance. We report the balanced accuracy on all our test sets for the SC data set in Table~\ref{tab:window_size}. For all these experiments, we use $20\%$ of the identities in the training set. The capacity of the \textbf{Supervised} model $C_{Sup} \circ E$ is independent of $W_{Self}$ (see sec.~\ref{sec:archi}), and thus so is its performance. We observe that the best performing models are the ones using $W_{Self}=2.5 sec$ for the Validation Set and $W_{Self}=5 sec$ for sets A and B. We argue that the features learned by the self-supervised model are less specific for larger window sizes. The PS operator drastically changes structured parts of the time series, but barely affects pure noise segments. As discussed in sec.~\ref{sec:phaseswap}, white noise is invariant with respect to the PS operator. With smaller window sizes, most of the segments are either noise or structured patterns, but as the window size grows, its content becomes a combination of the two. 

    \subsection{Frozen vs Fine-tuned Encoder}
    
    In Table~\ref{tab:summary}, we analyze the effect of freezing the weights of $E$ during the supervised fine-tuning. We compare the performance of the four variants described in sec.~\ref{sec:models} on the SC data set. All variants use $W_{Self}= 5 sec$, $W_{Sup}=30 sec$ and $r_{train}=20\%$. As expected, we observe that the \textbf{PhaseSwap} variant is the most performant one since it is less restricted in terms of training procedure than \textbf{PSFrozen} and \textbf{Random}. Moreover, the \textbf{PSFrozen} outperforms the \textbf{Random} variant on all test sets and is on par with the \textbf{Supervised} baseline on the Test set B. This confirms that the features learned during pre-training are useful for the downstream classification even when the encoder model $E$ is frozen during the fine-tuning. 
    The last variant, \textbf{Random}, allows us to disentangle the contribution of the self-supervised task from the prior imposed by the architecture choice for $E$. As we can see in Table~\ref{tab:summary}, the performance of the \textbf{PhaseSwap} variant is significantly higher than the latter variant, confirming that the self-supervised task chosen here is the main factor behind the performance gap.
    
    \begin{table}[t]
    \centering
    \caption{Balanced accuracy reported on the SC data set for the four training variants.}
    \label{tab:summary}
    \begin{tabular}{c@{\hspace{2em}}r@{\hspace{2em}}r@{\hspace{2em}}r}
    \toprule
    Experiment & Validation Set      & Test set A    & Test set B      \\ \midrule
    \textbf{Supervised} & 79.4\%          & 67.9\%        & 66.0\%            \\
    \textbf{PhaseSwap}  & \textbf{84.3\%} & \textbf{72.0\%} & \textbf{69.6\%} \\
    \textbf{PSFrozen}   & 75.2\%          & 68.1\%        & 67.1\%          \\
    \textbf{Random}     & 70.1\%          & 62.1\%        & 63.9\%          \\ \bottomrule
    \end{tabular}
    \end{table}
    
    \subsection{Architecture}
    Most of the experiments in this paper use the FCN architecture \cite{wang2017time}. In this section, we illustrate that the performance boost of the PhaseSwap method does not depend on the neural network architecture. To do so, we also analyze the performance of a deeper architecture in the form of the Residual Network (ResNet) proposed by Humayun \etal~\cite{humayun2019end}. We report in Table~\ref{tab:resnet-comparison} the balanced accuracy computed using the SC data set for two choices of $W_{Self} \in \{2.5 sec, 30 sec\}$ and two choices of $r_{train} \in \{ 20\% , 100\%^*\}$.  The table also contains the performance of the FCN model trained using the PS task as a reference. We do not report the results for the RP experiment using $W_{Self}=30sec$ as we did not manage to make the self-supervised pre-training converge. All ResNet models were trained for 15 epochs for the supervised fine-tuning.
    For $r_{train}=20\%$, we observe that pre-training the ResNet on the PS task outperforms both the supervised and RP pre-training. We also observe that for this setting, the model pre-trained with $W_{Self}=30sec$ performs better on both the validation set and test set B compared to the one pre-trained using $W_{Self}=5sec$. Nonetheless, the model using the simpler architecture still performs the best on those sets and is comparable to the best performing one on set A. We believe that the lower capacity of the FCN architecture prevents the learning of feature representations that are too specific to the pretext task compared the ones learned with the more powerful ResNet.
    For the setting $r_{train} = 100\%^*$, the supervised ResNet is on par with a model pre-trained on the PS task with $W_{Self}=30sec$. Recall that  $r_{train} = 100\%^*$ refers to the setting where all recording session and patients are used for the training set. Based on these results, we can conclude that there is a point of diminishing returns in terms of available data beyond which the self-supervised pre-training might even deteriorate the performance of the downstream classification tasks. 

    \begin{table}[t]
    \centering
    \caption{Evaluation of the \textbf{PhaseSwap} model using the ResNet architecture on the SC data set. Values denoted with a * are averages across two runs.}
    \label{tab:resnet-comparison}
    \begin{tabular}{c@{\hspace{1em}}c@{\hspace{1em}}c@{\hspace{1em}}c@{\hspace{1.5em}}c@{\hspace{1.5em}}c@{\hspace{1.5em}}c}
    \toprule
    $r_{train}$ & $W_{Self}$ & Architecture & Experiment & Val. Set       & Test set A      & Test set B      \\ \midrule
    20\%        & 5sec      & FCN          & FCN + PS   & 84.3\%           & 72.0\%          & 69.6\%          \\ \hline
    20\%        & 5sec      & ResNet       & phase swap  & 82.1\%           & \textbf{72.5\%} & 69.6\%           \\
    20\%        & 5sec      & ResNet       & RP         & 72.3\%           & 67.4\%          & 65.9\%          \\
    20\%        & -         & ResNet       & supervised & 79.1\%*          & 70.0\%*         & 66.5\%*         \\
    20\%        & 30sec     & ResNet       & phase swap  & \textbf{83.6\%}  & 70.7\%          & \textbf{69.3\%} \\ \hline
    100\%*      & 5sec      & FCN          & FCN + PS   & 84.3\%           & -               & -               \\ \hline
    100\%*      & 5sec      & ResNet       & phase swap  & 81.2\%           & -               & -               \\
    100\%*      & 5sec      & ResNet       & RP         & 79.1\%           & -               & -               \\
    100\%*      & -         & ResNet       & supervised & \textbf{84.2\%*} & -               & -               \\
    100\%*      & 30sec     & ResNet       & phase swap  & \textbf{84.2\%}  & -               & -               \\ \bottomrule
    \end{tabular}
    \end{table}
        
    \section{Conclusions}
    
    We have introduced the phase swap pretext task, a novel self-supervised learning approach suitable for bio-signals. This task aims to detect when bio-signals have mismatching phase and amplitude components. Since the phase and amplitude of white noise are uncorrelated, features learned with the phase swap task do not focus on noise patterns. Moreover, these features exploit signal patterns present both in the amplitude and phase domains. We have demonstrated the benefits of learning features from the phase component of bio-signals in several experiments and comparisons with competing methods. Most importantly, we find that pre-training a neural network with limited capacity on the phase swap task builds features with a strong generalization capability across subjects and observed sessions. One possible future extension of this work, as suggested by the results on the CHB-MIT data set \cite{shoeb2009application}, is to incorporate spatial correlations in the PS operator through the use of a spatio-temporal Fourier transformation.
    
    \subsubsection*{Acknowledgements}
    This research is supported by the Interfaculty Research Cooperation ``Decoding Sleep: From Neurons to Health \& Mind'' of the University of Bern.
    
	\bibliographystyle{splncs04}
	\bibliography{mybiblio}

\end{document}